\documentclass[10pt,twocolumn,letterpaper]{article}

\usepackage{wacv}              

\usepackage{graphicx}
\usepackage{amsmath}
\usepackage{amssymb}
\usepackage{booktabs}
\usepackage[accsupp]{axessibility}
\usepackage{algorithm}
\usepackage{algorithmic}
\usepackage{multirow}
\usepackage{booktabs}
\usepackage{float}
\usepackage{svg}

\usepackage[pagebackref,breaklinks,colorlinks]{hyperref}

\usepackage[capitalize]{cleveref}
\crefname{section}{Sec.}{Secs.}
\Crefname{section}{Section}{Sections}
\Crefname{table}{Table}{Tables}
\crefname{table}{Tab.}{Tabs.}

\def\wacvPaperID{643} 
\def\confName{WACV}
\def\confYear{2024}

\begin{document}

\title{Linking convolutional kernel size to generalization bias in face analysis CNNs}

\author{Hao Liang\\
Rice University\\
Houston, TX, USA\\
{\tt\small hl106@rice.edu}
\and
Josue Ortega Caro\\
Yale University\\
New Haven, CO, USA\\
{\tt\small josue.ortegacaro@yale.edu}
\and
Vikram Maheshri\\
Houston University\\
Houston, TX, USA\\
{\tt\small vmaheshri@uh.edu}
\and
Ankit B. Patel\\
Rice University\\
Houston, TX, USA\\
{\tt\small ankit.patel@rice.edu}
\and
Guha Balakrishnan\\
Rice University\\
Houston, TX, USA\\
{\tt\small guha@rice.edu}
}

\maketitle


\begin{abstract}
Training dataset biases are by far the most scrutinized factors when explaining algorithmic biases of neural networks. In contrast, hyperparameters related to the neural network architecture have largely been ignored even though different network parameterizations are known to induce different \emph{implicit biases} over learned features. For example, convolutional kernel size is known to affect the frequency content of features learned in CNNs. In this work, we present a causal framework for linking an architectural hyperparameter to out-of-distribution algorithmic bias. Our framework is experimental, in that we train several versions of a network with an intervention to a specific hyperparameter, and measure the resulting causal effect of this choice on performance bias when a particular out-of-distribution image perturbation is applied. In our experiments, we focused on measuring the causal relationship between convolutional kernel size and face analysis classification bias across different subpopulations (race/gender), with respect to high-frequency image details. We show that modifying kernel size, even in one layer of a CNN, changes the frequency content of learned features significantly across data subgroups leading to biased generalization performance even in the presence of a balanced dataset.  

\end{abstract}

\maketitle
\vspace{-1em}
\section{Introduction}

Algorithmic biases of a deep neural network, i.e., performance disparities across subgroups in the data distribution, are most often attributed to sampling biases in its training dataset where some groups of the data distribution have significantly lower or higher proportions than others. While an imbalance in the training data certainly has a strong influence on a deep network's algorithmic bias~\cite{albiero2020analysis,zietlow2022leveling,ramaswamy2021fair,du2020fairness,kearns2018preventing,torralba2011unbiased}, it is just one factor in the development pipeline. Examples of other important factors in a model's design include its parameterization and training objective function. While several previous works\cite{beutel2019fairness,wick2019,padala2020fnnc,lohaus2020, kleindessner_ordinal_regression, risser2020tackling} explore the impact of objective functions on bias and suggest fairness regularizers to include during training, a network's parameterization, i.e., its particular functional form, has been virtually unexplored in the context of bias. A neural network has several parameterization hyperparameters that must be set by its designer, including the number of layers, activation functions, and convolutional kernel sizes in the case of a convolutional neural network (CNN). Each of these choices can affect the type of features learned by the model, which could in turn impact bias. In this work, we take a first step in showing how to causally link a convolutional neural network's  kernel size to its algorithmic bias. 

Different network parameterizations are known to induce different \emph{implicit biases} over learned features. For example, CNNs tend to learn representations that are sensitive to high spatial frequency features of the input signal ~\cite{caro2020local,yin2019fourier,wang2020high}. Based on the Fourier uncertainty principle, this phenomenon may be attributed to the size of the convolutional kernels \cite{caro2020local} -- smaller kernel sizes result in features that span a greater range in the Fourier domain. This implicit feature bias may be exposed by injecting a certain high-frequency signal into test images and observing a drop in the algorithm's performance \cite{wang2020towards}. Implicit biases typically do not harm the model on within-distribution test samples because the parameters are well-tuned to the statistics in that distribution. The frequency noise/energy injection is essential to push the images \emph{out-of-distribution (OOD)}, thereby linking implicit biases to the \emph{generalization performance} of the network. In this work, we show that in addition to this effect, implicit biases also affect the algorithmic bias of a network, such that features used by the network for one data subgroup may have significantly different characteristics for another. This in turn, can lead to disparate OOD performance on these groups. 
 
The main contribution of our work is a causal framework for linking an architectural hyperparameter of a neural network to its OOD bias. Our framework is based on an \emph{experimental} procedure in which one or more parameters of a neural network architecture are modified at a time and the resulting bias on OOD samples are studied. First, we train from scratch multiple versions of the same network model that differ only in their choice of hyperparameter, e.g., convolutional kernel size. Second, we construct an OOD test set of images by injecting controlled perturbations to them, and obtain each model's prediction on each image. The perturbation type should be based on the intended implicit feature bias being studied. For example, in our experiments, we use adversarial attacks and energy injections in Fourier passbands to probe frequency biases. Third, we fit a linear regressor to predict a model's performance on an OOD test image as a function of the hyperparameter choice, degree of perturbation to the image, and various image attributes. Fourth and finally, we use the regression coefficients to measure the hyperparameter's causal effects on model performances across data subgroups. This analysis provides a quantitative answer to whether the hyperparameter has a disparate causal effect across data subgroups.

While our framework is general, we focused our experiments on studying the causal relationship between sensitivity to high-frequency image details induced by changes to convolutional kernel sizes and performance of face analysis classifiers across subpopulations (race/gender protected groups). We trained several research-grade face gender classifiers on public datasets, and show that modifying kernel size from a commonly used  range: $3\times 3$ to $11\times 11$ even in just the first layer of these CNNs will not only change the frequency content of learned features, but that this change can vary significantly across race/gender groups. We established this effect using both adversarial perturbations and energy injections to the high-frequency bands of the test images. This work opens the door to further careful studies on understanding the impact of neural network design decisions on algorithmic bias.

\section{Related Work}

\subsection{Fairness in computer vision}
Studies on fairness in computer vision predominantly focus on measuring and mitigating possible biases of computer vision models and datasets~\cite{albiero2020analysis,grother2019face,zietlow2022leveling,ramaswamy2021fair,quadrianto2019discovering,du2020fairness,kearns2018preventing,torralba2011unbiased}. Biases may be measured with a number of metrics~\cite{vasudevan2020lift,hardt2016equality,corbett2018measure} that quantify disparate performance differences of algorithms across population subgroups. Face recognition and analysis systems are often under the most scrutiny due to their sensitive nature~\cite{balakrishnan2021towards, karkkainen2021fairface,zietlow2022leveling}. Perhaps the most famous of these studies was ``Gender Shades'' study~\cite{gendershades}, which identified the systematic failings of face analysis systems on particular racial and gender demographics.

Natural image datasets are known to have sampling biases~\cite{albiero2020analysis,ponce2006dataset, torralba2011unbiased}, i.e., their joint distributions of attributes are far from random. For example, the CelebA face dataset is known to have a higher proportion of females with young ages compared to males~\cite{balakrishnan2021towards}. A model trained on such a dataset can inherit its biases~\cite{karkkainen2021fairface,zemel2013learning,ponce2006dataset}. Therefore, algorithmic fairness issues can be greatly mitigated if the algorithm is trained on a more balanced dataset. Human face datasets have been particularly scrutinized~\cite{balakrishnan2021towards,klare2012face,kortylewski2019analyzing,kortylewski2018empirically,merler2019diversity} as models trained on these data can exhibit systematic failings with respect to attributes protected by the law~\cite{kleinberg2018discrimination}. Multiple approaches to mitigate dataset bias include collecting more diverse examples~\cite{merler2019diversity}, using image synthesis to compensate for distribution gaps~\cite{balakrishnan2021towards, kortylewski2019analyzing, sattigeri2019fairness, singh2021matched, wang2020fair, zietlow2022leveling}, and resampling~\cite{li2019repair}. Our work, in contrast, is focused on understanding biases of deep learning models due to parameterization decisions instead of data. In addition, ~\cite{cruz2021promoting}, ~\cite{perrone2021fair}, ~\cite{rodolfa2021empirical},~\cite{baldini2021your} and~\cite{sellam2021multiberts} propose related ideas on searching for optimal hyperparameters taking fairness into account, but were not focused on computer vision tasks and architectures.

\subsection{Adversarial attacks}
 An adversarial attack perturbs an image until a given network changes its prediction, usually by applying gradient descent on the image. The resulting changes to the image are high frequency, and imperceptible to the human eye. This lack of robustness has sparked many theories~\cite{gilmer2018adversarial, mahloujifar2019curse,tanay2016boundary,ford2019adversarial,fawzi2018adversarial,bubeck2018adversarial,goodfellow2014explaining,schmidt2018adversarially,ilyas2019adversarial}. Recent work has shown that commonly found adversarial examples for state-of-the-art convolutional neural networks contain dataset-specific information \cite{wang2020high}. Furthermore, these adversarial attacks reflect properties of the features learned by the model \cite{caro2020local}, and that the model is biased towards certain features based on their architectural choice \cite{faghri2021bridging}. In this work, we analyze the information carried out by the attacks as a function of different architectural hyperparameter choices. Furthermore, we explore a novel hypothesis that adversarial attacks may allow us to expose differentiable information captured by a model's features across different dataset subpopulations.

\subsection{Frequency biases in CNNs}
In image processing, the most common way to represent pixel location is in the spatial domain by column (x), row (y), and z (value). The frequency (or Fourier) domain offers an alternative perspective on the signal, by decomposing it in terms of sinusoids of varying frequencies.

Several recent works have provided new insights into the behavior of CNNs by studying the relationship between frequency content in input signals and a CNN's predictions. For example, one finding is that high-frequency components play a significantly higher role in a CNN’s decision function and performance compared to human perception\cite{wang2020high}. Please refer to Figure \ref{fig:injection_example} for example of high/low-frequency components of an image.
Another study showed that a contributing factor to this is that convolutional operations in CNNs will introduce an \emph{implicit bias} towards using higher frequencies in an image~\cite{caro2020local}. Nonlinear activation functions such as the rectified linear unit (ReLU) could also be a contributing factor~\cite{karantzas2022understanding}. We build on the findings in these works to study the effect of frequency-based features in differentiable algorithmic performance across dataset subgroups like gender and race. The Discrete Fourier Transform (DFT)is commonly used to transform an image between the spatial and frequency domain.

\section{Methods}
\begin{figure*}[!t]
    \centering
    
    {\label{fig:1}\includegraphics[width=0.9\textwidth]{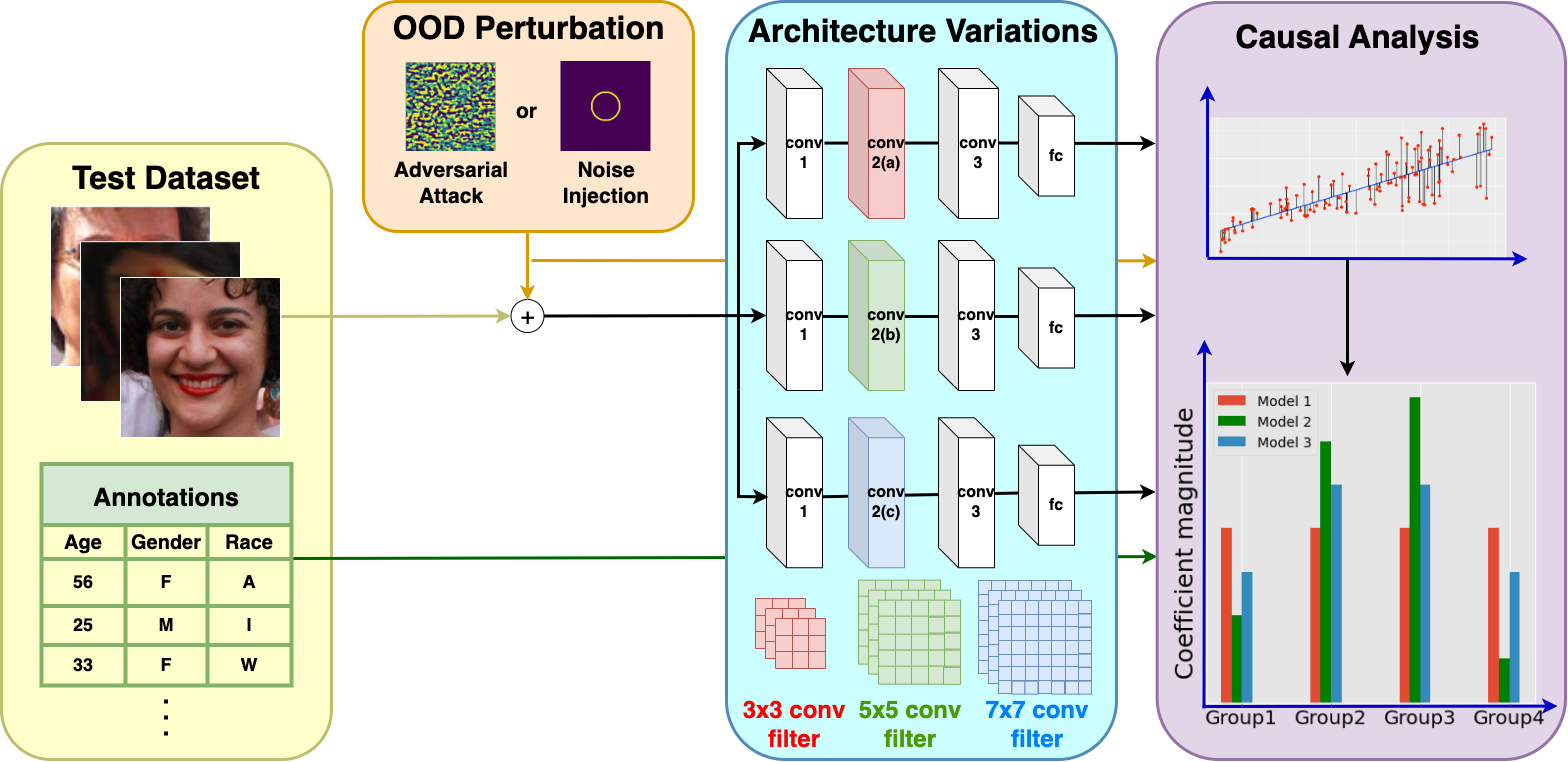}}
    \caption{\textbf{Method Framework}. The framework of our method consists of three parts. First, we perturb all the test images with Out Of Distribution(OOD) perturbations (in this work, we use adversarial attack perturbation and frequency noise injection), to make a new test dataset which contains the same images but with different noises injected. Second, we send the new test images to different models, where the models share most of the architecture design but differs only in some small part (e.g. convolutional kernel size). Last, we collect the results got from the last step and split them into according to sensitive attributes, and apply our causal analysis on the results.}
    \label{fig:frame2}
\end{figure*}

Our goal is to uncover the causal effect of convolutional kernel size on potential algorithmic biases due to an alteration of learned feature characteristics. We propose a framework to do this (see Fig.~\ref{fig:frame2} for an overview). Our framework consists of three key steps. We first train $K$ versions of the same network architecture that differ only by the choice of a single hyperparameter. The hyperparameter choice acts as a causal intervention, giving us an experimental rather than observational testing procedure. We then apply an out-of-distribution (OOD) perturbation to a set of test images that are annotated with various attributes of interest, including ``protected'' attributes (e.g., race and gender for faces) that we will use for bias analysis. Finally, we use a linear regression to predict some measure related to model predictions, given covariates such as the hyperparameter choice, image attributes, and OOD perturbation degree. We use the regression coefficients as estimates of causal effects of the various factors on the model, and specifically compare the differences between coefficients corresponding to \emph{protected} attributes to evaluate bias. We describe the three steps of our framework in the following sections.

\vspace{-1em}
\subsection{Architecture training}
We first train $K$ different versions of the same architecture, all identical in structure except for a modification to the hyperparameter of interest. We train all architectures on the same training dataset. We also initialize the weights and biases of all networks from identical normal distributions (i.e., identical mean and variances). After the networks are trained, we ``freeze'' their parameters, and will not modify them further in our framework.
\vspace{-0.5em}
\subsection{OOD perturbations to test data}
\label{sec:ood}
Given various trained architectures, our goal is to amplify biases across their learned features. One option is to run test samples coming from the same distribution as the training data through these networks, and measure performance across different protected attribute subgroups. The problem with this strategy is that deep neural networks are over-parameterized and therefore able to fit any training distribution nearly perfectly. Hence, even if a hyperparameter is altered from one network to another, both networks will likely yield similar performances on training data points. 

However, as demonstrated in past works \cite{wang2020towards,yin2019fourier}, out-of-distribution (OOD) samples can paint a far different picture, with some models suffering in performance compared to others, thereby exposing differences across learned features. Therefore, a key step in our model is to inject a test set of images with a subtle class of perturbations so that they become OOD. In our experiments, we focus on frequency-related implicit biases of CNNs, and so we consider two types of perturbations from the neural network literature: adversarial attacks, and frequency energy injections.

\vspace{-0.4cm}
\subsubsection{Adversarial attacks}
\label{sec:aa_method}
We consider two types of adversarial attacks in our experiments. The first, \textbf{FGSM}\cite{goodfellow2014explaining}, applies gradient descent on the loss of the network's output with respect to the input image to ``nudge'' the image in incremental steps towards a direction that changes the network's prediction. The second, \textbf{CW attack}\cite{carlini2017towards}, utilizes two separate losses: a gradient-based loss to make the classifier change its prediction (similar to FGSM), and a regularization to make the magnitude of the change to the image as small as possible. This makes the perturbation distance(i.e. $l_2$ norm of the difference between perturbed image and original image) of CW attack a useful metric for measuring the degree of difficulty to perturb an image. We show an example of a CW and FGSM attack for the same input image in Figure 1 in Supplementary, which further shows that CW perturbation is an order of magnitude smaller due to the effect of its regularization.

\vspace{-0.4cm}
\subsubsection{Frequency energy injection}
\label{Sec:noise injection}
\begin{figure*}[t!]
    \centering
    {\label{fig:3}\includegraphics[width=\textwidth]{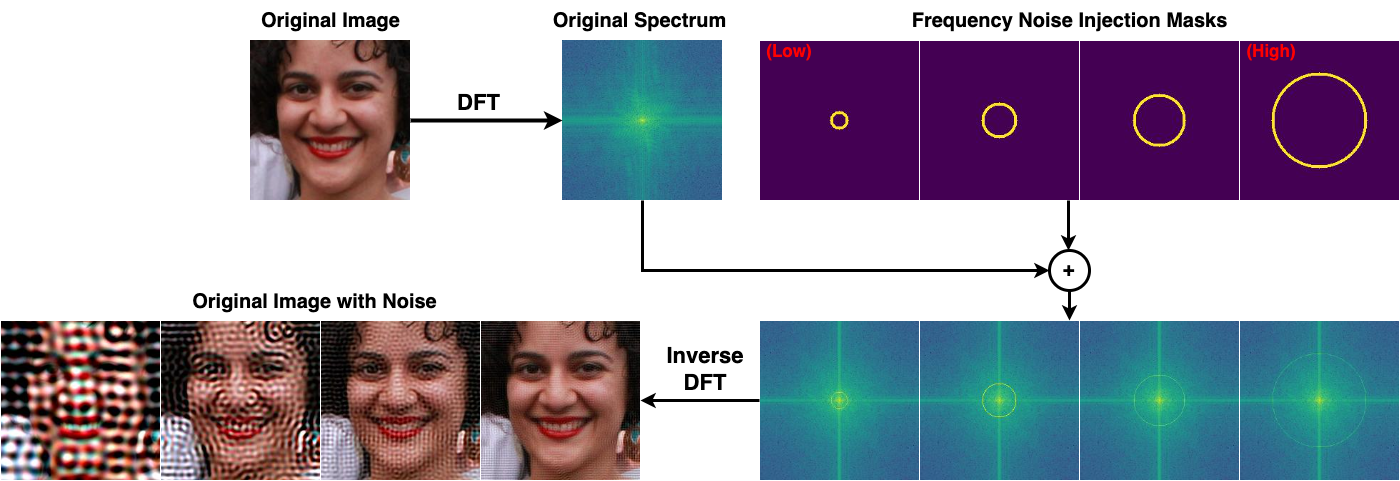}}
    
    \caption{\textbf{Examples of frequency energy injection perturbations}. Our goal with this perturbation is 'jitter' all frequencies with the same magnitude in an input image. To do this, we inject random noise to the Fourier coefficients of an image lying on an annulus of a particular radius in the Fourier spectrum (top row of annulus), according to: $(1+\delta)|A|e^{-j\phi}$.
 }   

    \label{fig:injection_example}
\vspace{-1em}
\end{figure*}

We also experiment with injecting energy to a specific frequency band to obtain a more fine-grained link between frequency content and network features. Fig.~\ref{fig:injection_example} depicts our process. For each test image, we use the DFT to obtain a Fourier spectrum, and amplify the amplitudes of Fourier coefficients lying on an annulus in the spectrum. In particular, let $F[\omega_x, \omega_y] = |A|e^{-j\phi}$ represent a complex coefficient in the Fourier spectrum of an image at location $(\omega_x, \omega_y)$ (corresponding to $x$ and $y$ frequencies), with radius $r = \sqrt{\omega^2_x + \omega^2_y}$, lying in the annulus defined by $(r - r_0)^2 \leq \Delta^2$. We increase the amplitude $A$ by a factor of $1+\delta$, to yield a modified coefficient $F'[\omega_x, \omega_y]=(1+\delta)|A|e^{-j\phi}$. In our experiments, we set $\Delta=2$, and $\delta=15$ and $r_0>0$ is the frequency radius into which we are injecting energy~\cite{yin2019fourier}. If $r_0$ is small (large), we are modifying low (high) frequency components of the image. Finally, we reconstruct the perturbed image using an inverse DFT.

\vspace{-0.4em}
\subsection{Causal analysis}
\label{sec:causal}
\vspace{-0.5em}

We run the test set of OOD images through the $K$ networks, yielding $K$ predictions per image. We assume each image also comes with annotations for various relevant semantic attributes (including protected attributes with which we may compute algorithmic bias measures), as well as perturbation attributes (e.g., frequency of energy injection).  Our goal is to measure the causal effects of the architectural hyperparameter of interest (kernel size) on model performance per protected attribute subgroup.  

To do so, we use a multivariable linear regression model that predicts a dependent variable from multiple independent variables. For test image $i$ processed in network $k$, let $x_k$ be the corresponding hyperparameter and $y_{ik}$ be a measure of network performance on image $i$. Then we can specify the following regression equation:
\begin{equation}
\label{eq:reg1}
    y_{ik} = \beta x_k + \epsilon^0_{ik},
\end{equation}
where $\epsilon_{ik}$ is an error term. Our coefficient of interest is $\beta$. Under the assumption that $E[x_k \cdot \epsilon^0_{ik}]=0$, we can interpret $\beta$ as the causal effect of network architecture on performance. Of course, this independence assumption is unlikely to hold, as image attributes, including the OOD perturbation value, will generally affect a neural network's performance. 

We can weaken this assumption using a vector of image attributes $\boldsymbol{Z_i}$, and augmenting equation \eqref{eq:reg1} as follows:

\begin{equation}
\label{eq:reg2}
    y_{ik} = \beta x_k + \boldsymbol{Z_i}^\prime \boldsymbol{\gamma} + \epsilon^1_{ik},
\end{equation}

where $\boldsymbol{\gamma}$ is a vector of coefficients, and $\beta$ is the causal effect of network architecture on performance under the weaker assumption $E[x_k \cdot \epsilon^1_{ik} | \boldsymbol{Z_i}]=0$. Moreover, we hypothesize that the effect of architecture hyperparameter $x$ on performance may vary by protected attributes, a subset of all image attributes in $\mathbf{Z_i}$. In order to allow for this possibility, we further augment equation \eqref{eq:reg2} as follows:

\begin{equation}
\label{eq:reg3}
    y_{ik} = \boldsymbol{P_i}^\prime \boldsymbol{\beta} x_k + \boldsymbol{Z_i}^\prime \boldsymbol{\gamma} + \epsilon^1_{ik},
\end{equation}

where $\boldsymbol{P_i}$ is a vector of protected image attributes, and $\boldsymbol{\beta}$ is now a vector of coefficients. We use a heuristic approach to choose the vectors $\boldsymbol{Z_i}$ and $\boldsymbol{P_i}$ that is commonly used for causal inference in the social sciences~\cite{wooldridge2010econometric}. First, we incrementally add controls to $\boldsymbol{Z_i}$ and test whether our estimates of $\boldsymbol{\beta}$ change under alternative specifications (using an F-test with the null-hypothesis that the estimates of $\boldsymbol{\beta}$ are equal across specifications). This is a test of the exogeneity assumption; if $\boldsymbol{Z_i}$ is a sufficiently rich vector of controls to satisfy $E[x_k \cdot \epsilon^0_{ik} | \boldsymbol{Z_i}]=0$, then the assumption will also be satisfied conditional on an augmented vector of controls. 
Second, we start with a rich vector $\boldsymbol{P_i}$ to allow for the effect of network architecture on performance to be highly flexibly estimated. In our application, we begin by specifying $\boldsymbol{P_i}$ as a fully saturated vector of dummy variables corresponding to all protected attribute combinations (e.g., White Male, White Female, etc.) and estimate $\boldsymbol{\beta}$. We then test whether the elements of $\boldsymbol{\beta}$ are equal to each other using pairwise F-tests. 
If we are unable to reject equality of coefficients, we cannot reject that the effect of network architecture on performance varies across those two groups.
 

$\boldsymbol{\beta}$ encodes the joint causal effects of hyperparameter value $x$ and protected attributes in $\boldsymbol{P_i}$ on output $y$. In particular, $\boldsymbol{\beta}_g$ is the expected change in $y$ due to a unit change to $x$, when feature $g$ is ``True'' (set to 1). In our experiments, we compare the values in $\boldsymbol{\beta}_g$ corresponding to different protected attribute subgroups to one another (see Table 2 in Supplementary, and Fig.~\ref{fig:regress_inject}).

\vspace{-0.4cm}
\section{Experiments \& Results}
\vspace{-0.15cm}
We evaluated our work on the task of gender classification from face images using two popular datasets: Fairface \cite{karkkainen2019fairface} and UTKFace \cite{zhifei2017cvpr}. Fairface contains a roughly equal number of samples from different race/gender groups, and has 86,744 training and 10,954 testing samples. FairFace contains labels for 7 race groups (`East Asian', `White', `Latino Hispanic', `Southeast Asian', `Black', `Indian', `Middle Eastern') and 2 gender groups (`Male' and `Female'). UTKFace contains 20,000 training and 3,705 testing samples, but is not balanced across race  groups. It contains labels for 5 race groups (`White', `Black', `Asian,' `Indian', `Others') and two gender groups ('Male' and 'Female'). We remove faces from `Others' because they have no consistent characteristics. To mitigate effects of sampling biases during training, we used inverse sampling based on the number of examples from each race group. Training details are in Section A in Supplementary.

We demonstrate results using the ResNet-34~\cite{he2016deep} base architecture for our experiments but obtained similar results using two other popular networks: DenseNet~\cite{huang2017densely} and VGG-16~\cite{simonyan2014very}. Please refer to Section D, E in  Supplementary for results using these two models. Our architectural hyperparameter of interest was convolutional kernel size. We considered two different scenarios: changing only the kernel size of the first layer and changing the kernel size of all layers simultaneously. Interestingly, both scenarios yielded similar results, and so we leave results for the latter in Figure 6 in Supplementary. We varied the first layer kernel size (FLKS) within the range $[3, 11]$, which encompasses the popular choices for this hyperparameter for nearly all CNNs in the literature. We initialize the weights and biases of all of our models randomly by drawing from a Normal distribution with variance set to $0.02$. For each network and kernel size value, we trained 3 independent models and presented average results to mitigate the influence of random initialization factors.

We report our networks' accuracies for different race groups on \emph{non-OOD} test images in Table 1 in Supplementary, to demonstrate that they all achieve high accuracies on both datasets. The performances do not significantly vary with FLKS because the training and testing images are all from the same distribution. We now present our results separately for the two OOD perturbations described in Sec.~\ref{sec:ood}: adversarial attacks and frequency energy injections.

\begin{figure*}[!t]
    \centering
    {\label{fig:1}\includegraphics[width=0.9\textwidth]{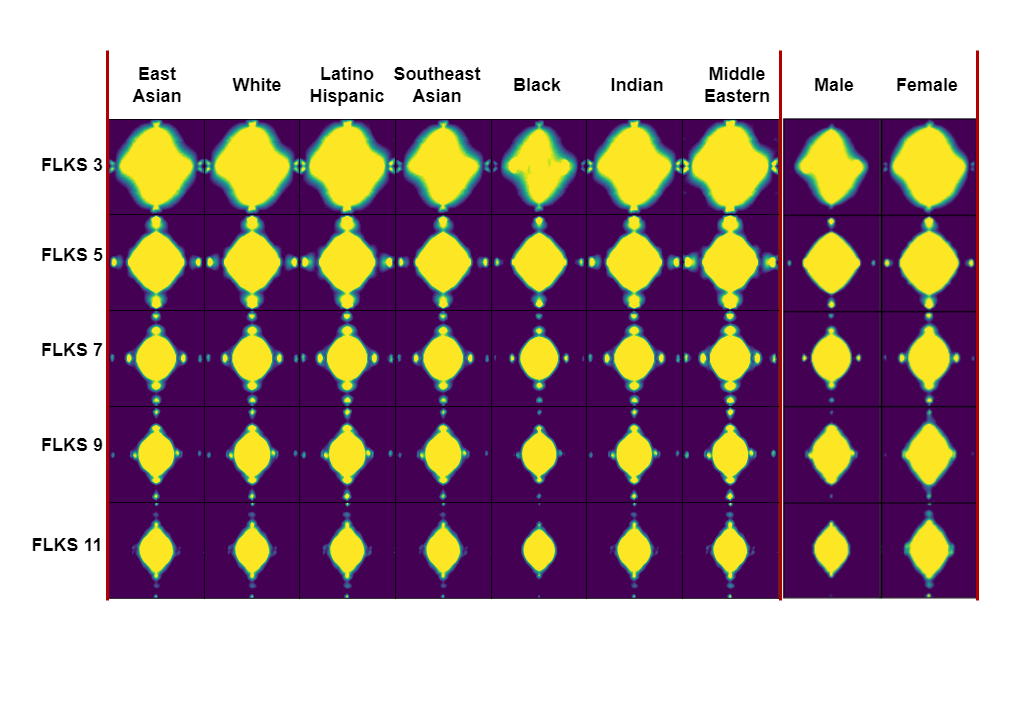}}
    \vspace{-6em}
    \caption{\textbf{Average spectra of adversarial perturbation images split by race and gender for Fairface}. Each row represents a model with a different first layer kernel size (FLKS). As FLKS increases, the spectra become more concentrated at low frequencies. The spectra for the Black race group consistently have less energy at high frequencies compared to the spectra of other race groups. Male spectra also have lower high frequency information compared to Female spectra. These results demonstrate that changes to FLKS induce different feature biases for networks, which also vary by protected attribute subgroups. See Figure 7 in Supplementary for the analogous spectra for the UTKFace dataset.}
    \vspace{-1em}
    \label{fig:spectra_fair}
\end{figure*}

\vspace{-0.3cm}
\subsection{Adversarial attacks}
\label{exp:aa}
We present results in this section using the CW adversarial attack. We obtained similar results using FGSM (see Figure 8 in Supplementary).
\vspace{-0.5em}

\subsubsection{Analyzing Fourier spectra}
\label{sec:spectra}
We first visualize the average Fourier spectra magnitudes of the adversarial perturbation images split by race/gender groups and FLKS in Fig.~\ref{fig:spectra_fair} (for Fairface) and the results for UTKFace are in Figure 7 in Supplementary. Results on both datasets show similar trends. First, as FLKS increases, the spectral energy becomes more focused at low-frequencies (closer to center). Second, holding FLKS value constant, we see that the spectrum for the Black group consistently contains less high-frequency energy compared to the spectra of other race groups. This result also holds for the Male group compared to Female. The difference between different subgroups shrinks as FLKS increases, in line with findings from a previous study showing that low FLKS leads to higher implicit frequency bias~\cite{caro2020local}.

\begin{figure*}[!hbtp]
    \centering
    
    {\label{fig:1}\includegraphics[width=0.8\textwidth]{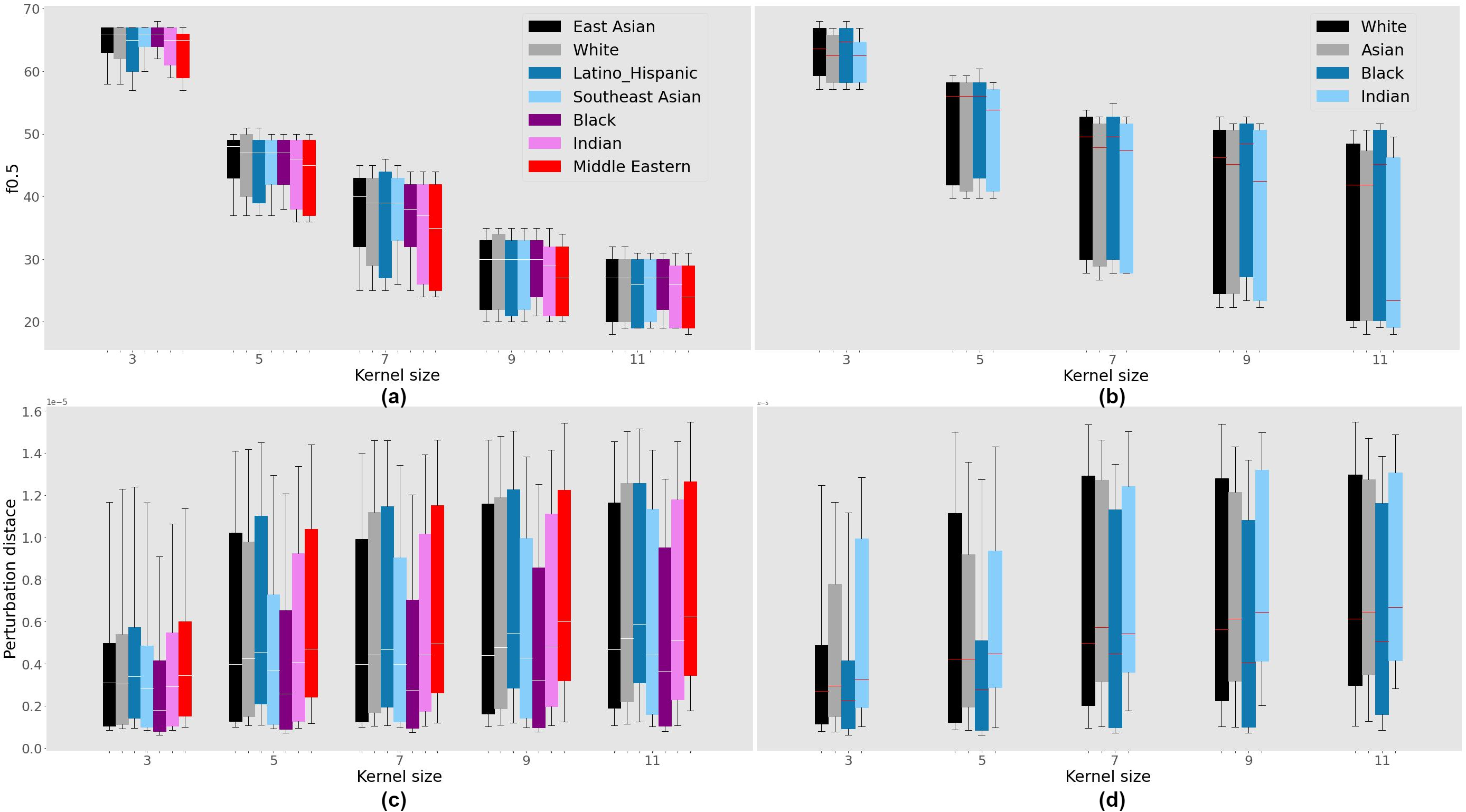}}
    \caption{\textbf{$f_{0.5}$ measures for adversarial perturbation spectra \& adversarial perturbation distance} Each boxplot shows the median score (white/red line in the boxes) and the $15\% - 85\%$ confidence interval for a different protected attribute group. The x-axis indicates the models' first layer convolutional kernel size (FLKS). {\bf (a) and (b)} are $f_{0.5}$ measures for adversarial perturbations using Fairface and UTKFace, respectively. The  $f_{0.5}$ score drops as FLKS increases for all demographic groups, which indicates that the adversarial attack focuses less on high-frequency information of the image for larger FLKS. {\bf (c) and (d)} show the adversarial perturbation distances per race group using Fairface and UTKFace, respectively, where distance is simply the $l_2$ norm of the perturbation image. As the FLKS increases, the perturbation distances generally increase too for all the demographic groups. In addition, for each FLKS value, the perturbation distances for the Black group are always significantly lower than those for other demographic groups.}
    \label{fig:f0.5}
\end{figure*}

To quantitatively assess differences in the perturbation spectra, we also compute the $f_{0.5}$ metric, known as ``half power frequency,'' or the frequency below which half of the signal's power lies. $f_{0.5}$ is a robust measure of energy concentration in a spectrum. Fig \ref{fig:f0.5}-top shows the $f_{0.5}$ scores for the spectra, confirming the visual trend observed in Fig.~\ref{fig:spectra_fair}. Please refer to the caption of Fig.~\ref{fig:f0.5} for more details.

\begin{figure*}[!hbtp]
    \centering
    
    {\label{fig:4}\includegraphics[width= 0.8\textwidth]{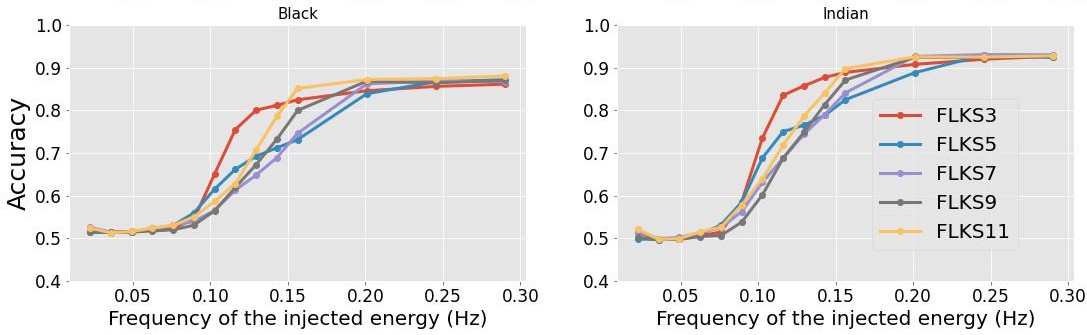}}
    \caption{\textbf{Frequency energy injection result}. We show models' performances with different FLKS for Black and Indian race groups separately (Please refer to Figure 9 in Supplementary for results of other race groups). In each individual figure, the x-axis is the frequency we are injecting energy at and the y-axis is the accuracy of different models. It is obvious that all the models suffer from low to mid frequency's energy injections, and become robust to mid to high frequency noises. It is hard to directly tell which group is getting influenced more than the others, which furthers asks for a quantitatively analysis.}
    \label{fig:res_injection} 
\vspace{-1em}
\end{figure*}

\begin{figure*}[t!]
    \centering
    
    {\label{fig:4}\includegraphics[width=0.8\textwidth]{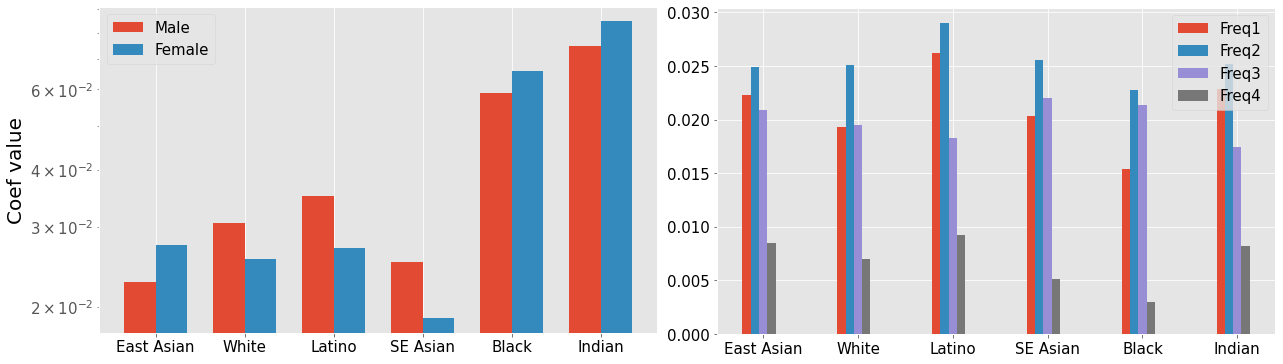}}
    
    \caption{\textbf{Regression coefficient values for $\boldsymbol{\beta}$ for different race groups. (a)} $\boldsymbol{\beta}$ values for the regression in Sec.~\ref{sec:causal-dist} linking kernel size to adversarial perturbation distance, also shown in Table 2 in Supplementary. There are significant differences across protected groups, e.g., the Black and Indian group has significant higher values compared to the other groups. {\bf (b)} $\boldsymbol{\beta}$ values for the regression in Sec.~\ref{sec:causal_acc}. The coefficients for Black are always lower than those of other race groups. In addition, the coefficient values for different frequencies within the same race group are also significantly different.}
    \label{fig:regress_inject} 
\vspace{-1em}
\end{figure*}

\vspace{-0.5cm}
\subsubsection{Perturbation distances}
\label{sec:ad_dist}
\vspace{-0.5em}
We next present the average perturbation \emph{distances} of adversarial attacks across race groups and models in Fig \ref{fig:f0.5}-bottom. The perturbation distance $d_p$ between an original test image $I$ and the perturbed image $I'$ may be computed by simply taking an L2 norm: $d_p(I, I') = ||I' - I||_2$, and quantifies how close/far the perturbed image to the original image. A larger distance indicates that more ``work'' must be done harder to fool the model, and its a reflection of the robustness of the model to other OOD perturbations \cite{li2022robust}. 

Results show that perturbation distance (and variance) increases with FLKS for all race groups. It is therefore harder to adversarially attack a model with a larger FLKS, likely because such a model focuses more of its energy on low-frequency image information (see Fig \ref{fig:spectra_fair}) and is therefore robust. In addition, we see that $d_p$ for the Black group is significantly lower than that of other groups. Please refer to the caption of Fig \ref{fig:f0.5} for more details.

\vspace{-0.8em}
\subsubsection{Causal analysis}
\vspace{-0.5em}
\label{sec:causal-dist}
Next, we quantitatively analyze the causal relationship between race and gender on perturbation distance $d_p$ by applying our causal analysis framework introduced in Sec. \ref{sec:causal}. Specifically, using Eq. \ref{eq:reg3}, we set $y$ to be $d_p$, and set both $\mathbf{P_i}$ and $\mathbf{Z_i}$ to contain ``dummy variables'' corresponding to all race/gender combinations. We use the Fairface dataset for this analysis, and use the race groups East Asian, White, Latino Hispanic, Southeast Asian, Indian and Black, and gender groups of Male and Female.

We use the \textit{statsmodel} package from Python to run this regression, and the results of values of $\beta$ and $\gamma$ are in Table 2 in Supplementary. The results of $\beta$ are also shown in Fig.~\ref{fig:regress_inject}(a). Based on the results, it is obvious that the coefficients for Black and Indian are significantly higher than that of other race groups, indicating the impact of kernel size on perturbation distance is much more significant for the two groups. White, Black and Indian female have larger $\beta$ values than their corresponding male group.

\subsection{Frequency Energy Injection}
\label{sec:res noise}
We next perform frequency-based OOD perturbations to the test images as described in Sec.~\ref{Sec:noise injection} and visualize results in Fig.~\ref{fig:res_injection}. Accuracies of all models/groups are more influenced by perturbations to low-to-mid frequencies ($0.02 -- 0.20 (Hz)$) than to mid-to-high frequencies. FLKS of 3 is less affected by frequency injections in the range $(0.08 - 0.15 (Hz))$. However, in general, it is difficult to distill significant trends from the plots alone. 

\vspace{-0.8em}
\subsubsection{Causal analysis}
\vspace{-0.5em}

\label{sec:causal_acc}

Similar to Sec.~\ref{sec:causal-dist}, we now perform a regression to measure the impact of kernel size, frequency of energy injection, and protected attribute subgroup on model error per image. Using Eq.~\ref{eq:reg3}, we set $y$ to be the error rate of an image and set both $\mathbf{P_i}$ and $\mathbf{Z_i}$ to contain ``dummy variables'' corresponding to race/frequency (of injected energy) combinations. We use four frequency subgroups: $\{ (0.05,0.07), (0.09,0.11), (0.13, 0.15), (0.17,0.19) (Hz)\}$, which we label $1,2,3,4$ for convenience. We report the results of coefficient $\beta$ values in Fig \ref{fig:regress_inject}(b). It is clear that the coefficients for frequency group 4 are significantly smaller than those of the rest of the frequencies, indicating that changes to kernel size influence the performance less on the OOD samples under relatively high-frequency injections. The coefficient for the Black group in frequency group 1 is also significantly smaller than those of the other groups. As the frequency increases, this gap reduces.

\vspace{-0.5em}
\section{Discussion and Conclusion} 

Our results in Figs. \ref{fig:spectra_fair}, \ref{fig:f0.5}, \ref{fig:res_injection}, \ref{fig:regress_inject} first demonstrate that smaller convolutional kernel sizes can cause a CNN to be biased towards high-frequency features, and increasing the kernel size mitigates this bias. We also see that such frequency bias significantly differed across different race/gender subgroups. All models trained on the two datasets focused less on high-frequency features for the Black and Male subgroup. While different features do not necessarily indicate performance bias on test samples, our results allow us to conclude that these differences do lead to performance biases on out-of-distribution (OOD) samples. We observed that this is the case for two different types of OOD image perturbation operators: adversarial attacks and frequency domain energy injections.

Different population subgroups will have different image characteristics. For example, the Black group will likely have darker skin tones than other race groups, and Females will have more hair on average than Males. Hence, it is not surprising that there is some difference in how images from one race are processed by a network compared to another. However, our results indicate something more significant: that there is a fundamental difference in the frequency characteristics of the image features across groups used by the network to make its decision. This difference may also lead to a performance bias, depending on the type of OOD data model is faced with. Our two OOD perturbations, while conceptually clear and well-motivated, are not associated with any real phenomena. It would be an interesting next step to relate frequency biases of features to disparate model performance on real-world OOD artifacts like shot noise, fog, and motion blur~\cite{yin2019fourier}.

Our work has several limitations. We cannot draw broad conclusions about the nature of kernel size for general CNNs across all applications, becaused we focused on a single application of interest. 
A further evaluation on a wider set of application domains is an important next step. We also limited our causal analyses to a few key variables. However, causal analysis typically relies on the ``no hidden counfounders'' assumption. An exhaustive set of image factors will help in computing more precise causal effects.

We focused on convolutional kernel size of a network in this work due to past results establishing a clear link between this hyperparameter and frequency content \cite{caro2020local}. However, our framework is agnostic to the nature of the hyperparameter. Indeed, next steps in this research space include similar analyses into a more comprehensive set of network hyperparameters, such as activation functions, depth of layers, weight initialization strategies, and even high-level designs (e.g., residual connections, transformer modules). We see our work as a first step in the important direction of understanding how neural network design choices impact bias, and hence, the fairness of these systems in our society.

\clearpage
{\small
\bibliographystyle{ieee_fullname}
\bibliography{egbib}
}

\onecolumn
\appendix
\begin{center}
\textbf{\large Supplemental Materials}
\end{center}

\section{Experiment setup}
\subsection{Training details}
The models were trained in a multi-task style. During training, the model was required to predict labels of race \& gender \& gender of a training sample, while during inference we only use the gender label predicted by the model. For training hyperparameters, we used a batch-size of 128, an initial learning rate of $1e^{-3}$ which decays by $10$ times at epoch 13 and epoch 17. We trained all the models with a total number of 21 epochs. We used an Adam optimizer.

\subsection{Computing resources}
All the experiments were run with NVIDIA Tesla A100 GPUs. Training a model per run took ~1 GPU hour. Applying FGSM attack to test dataset of FairFace took  $\sim 1,000$ seconds, applying CW attack to test dataset of FairFace took $\sim 6,000$ seconds. 

\section{Model performance}
We report our networks’ accuracies for different race groups on non-OOD test images in Table \ref{tab:acc}, to demonstrate that they all achieve reasonably high accuracies on both datasets. The performances do not significantly
vary with FLKS because the training and testing images are all from the same distribution. 

\begin{table}[H]
    \caption{{\bf Model performances on \emph{unperturbed} (non-OOD) images from the Fairface \& UTKFace datasets.} We report the trained models' performances on test sets split by race group.  Each number is an average over 3 different trained models. The first column indicates the dataset name and model's First Layer Kernel Size (FLKS). Fairface has 7 annotated race groups and UTKFace has 4. The performances are relatively constant with a variation to kernel size because the test and training images belong to the same distribution.}
    \label{tab:acc}
    \centering
    \begin{tabular}{|c|cccccccc|}
      \cline{1-9}
      \multicolumn{1}{|c|}{Dataset (FLKS)}&
      \multicolumn{1}{c|}{Overall} &
      \multicolumn{1}{c|}{White} &
      \multicolumn{1}{c|}{Black} &
      \multicolumn{1}{c|}{East Asian} &
      \multicolumn{1}{c|}{Indian} &
      \multicolumn{1}{c|}{Southeast Asian} &
      \multicolumn{1}{c|}{Latino} &
      \multicolumn{1}{c|}{Mid. Eastern}\\ \hline
      Fairface (3) &0.947 & 0.950& 0.894 & 0.942 & 0.945 & 0.894 & 0.957 & 0.977\\
      Fairface (5) &0.949  & 0.947 & 0.896 & 0.939 & 0.957 & 0.896 & 0.957 & 0.980 \\
      Fairface (7) &0.946  & 0.943 & 0.895 & 0.947& 0.956& 0.895 & 0.963 & 0.978 \\
      Fairface (9) &0.947  & 0.946 & 0.895 & 0.937 & 0.951& 0.895 & 0.960 & 0.979 \\
      Fairface (11) &0.946  & 0.949 & 0.892 & 0.937 & 0.949 & 0.885 & 0.967 & 0.979 \\ \hline
      UTKFace (3) &0.929   & 0.949 & 0.905 & 0.931 & 0.942 & /  & /  & / \\
      UTKFace (5) &0.935  & 0.951 & 0.901 & 0.940 & 0.951 & /  & /  & / \\
      UTKFace (7) &0.934  & 0.955 & 0.905 & 0.939 & 0.953 & /  & /  & / \\
      UTKFace (9) &0.937  & 0.955 & 0.910 & 0.941 & 0.955 & /  & /  & / \\
      UTKFace (11) &0.936 & 0.950 & 0.901 & 0.943 & 0.955 & /  & /  & / \\ \hline
    \end{tabular}
    
\end{table}

\clearpage

\section{Adversarial attack example}
 We show an example of a CW and FGSM attack for the same input image in the Supplementary, which further shows that CW perturbation is an order of magnitude smaller due to the effect of its regularization.
\begin{figure*}[h]
    \centering
    {\label{fig:1}\includegraphics[width=0.6\textwidth]{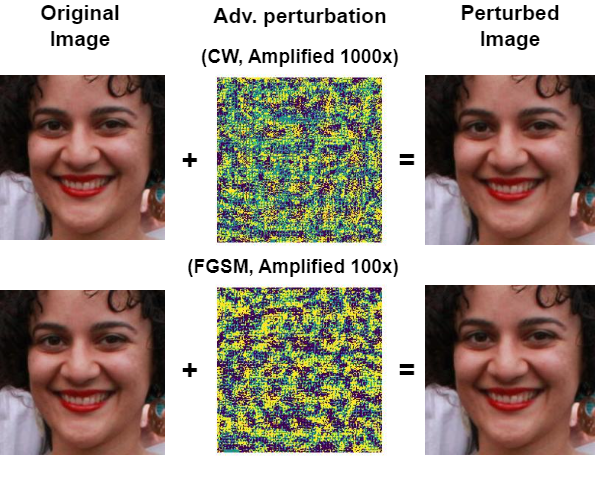}}

    \caption{\textbf{Example of adversarial attack perturbations}. By adding tiny noise-like perturbations (center image, amplified 100/1000 times for visualization) to a test image (left), a target neural network will output a wrong prediction. However, the perturbed image (right) has no perceptible differences with the original image to the human eye. We use the CW and FGSM attacks in our experiments.}
    \label{fig:aa_example}
\end{figure*}
\clearpage
\section{Results on DenseNet121}
\begin{figure*}[h]
    \centering
    
    {\label{fig:1}\includegraphics[width=0.8\textwidth]{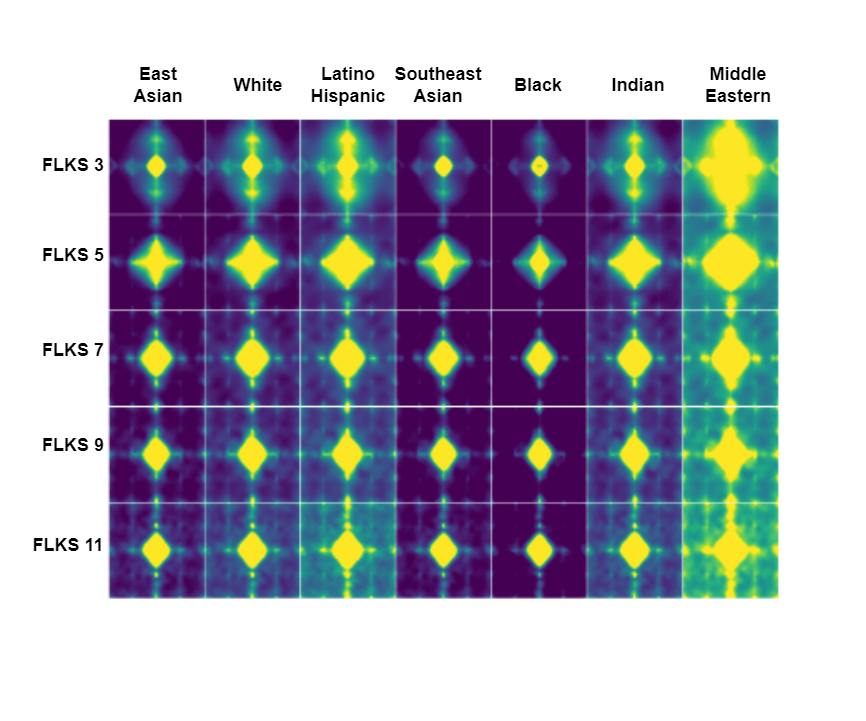}}
    \vspace{-2cm}
    \caption{\textbf{Perturbation Spectrum Visualization on Fairface using DenseNet121.} Similar to results in Figure 3 in main text, each row represents a model with a different First Layer Kernel Size(FLKS), and each column corresponds to protected attribute groups. We observed a similar trending as discovered in the other 2 results above: generally, the perturbation shifts it's attention to low-frequency information as FLKS increases, and the perturbations for Black always have lower high-frequency focus compared to other race group.}
    \label{fig:spectra_densenet}
\end{figure*}

\begin{figure*}[!hbtp]
    \centering
    
    {\label{fig:1}\includegraphics[width=\textwidth]{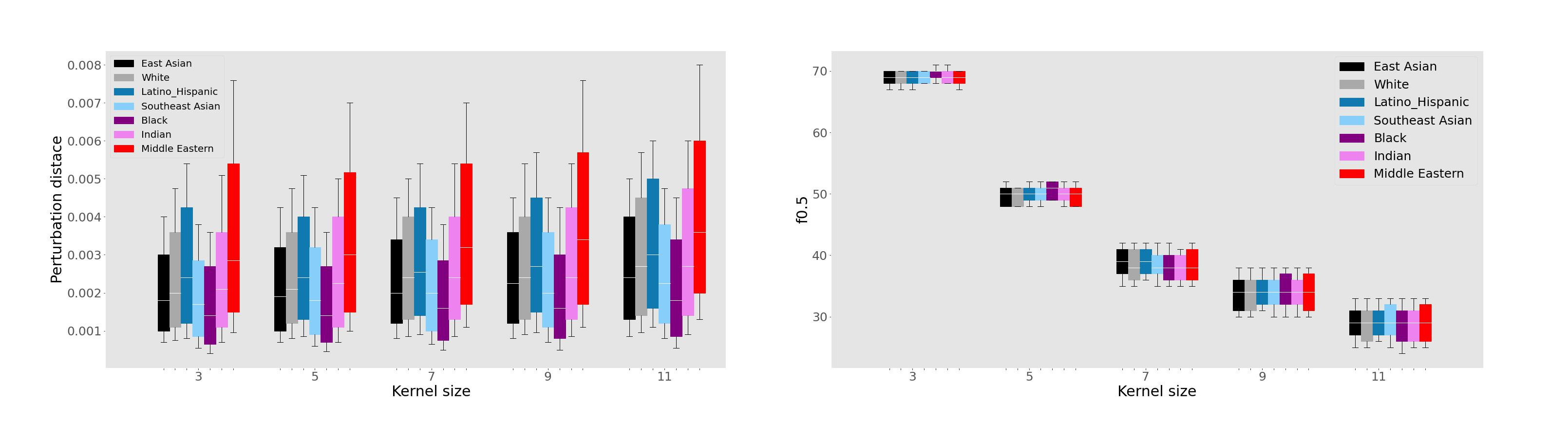}}
    \caption{\textbf{Perturbation spectra $f_{0.5}$ \& Perturbation distance for DenseNet121.} We visualize the perturbation and $f_{0.5}$ scores in the same way discussed in Sec. 4.1 and Figure 4. We observed a similar trend: there is a significant trend that the  $f_{0.5}$ drops as the FLKS increases for all demographic groups and as the FLKS increases, the perturbation distances generally increase too for all the demographic groups.}
    \label{fig:f0.5_densenet}
\end{figure*}

To further test the robustness and universality of our framework and conclusion, we also tested on DenseNet121 -- another popular face analysis model. We also vary the first convolutional kernel size from $\{3,5,7,,9,11\}$. We report the averaged spectra in Figure \ref{fig:spectra_densenet}, and its corresponding perturbation distance \& $f_{0.5}$ scores in Figure \ref{fig:f0.5_densenet}.

\clearpage
\section{Results on Vgg16}
\begin{figure*}[h]
    \centering
    
    {\label{fig:1}\includegraphics[width=0.8\textwidth]{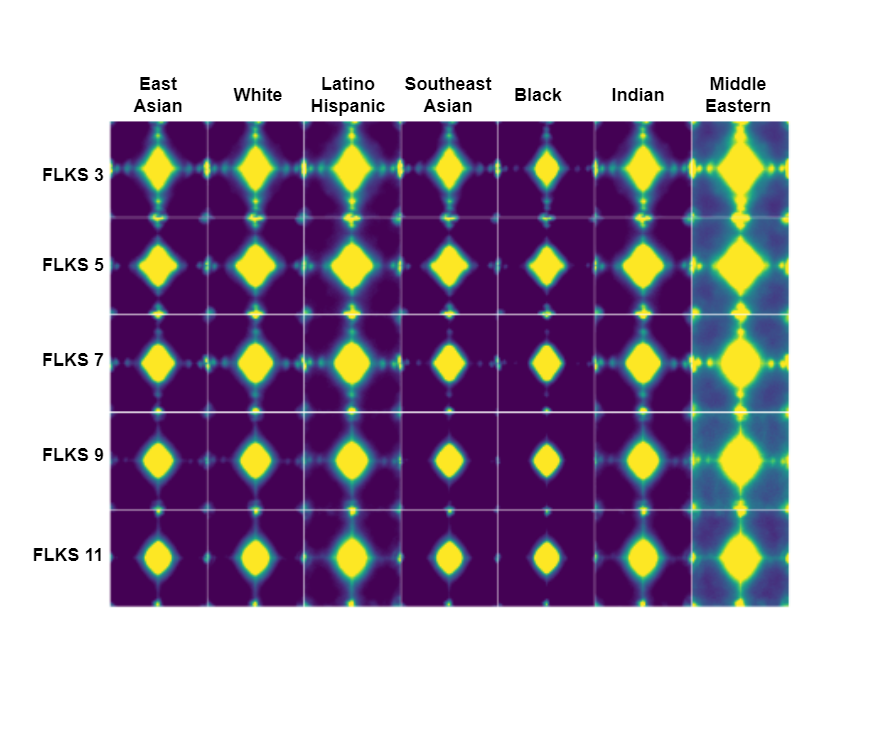}}
    \vspace{-2cm}
    \caption{\textbf{Perturbation Spectrum Visualization on Fairface using Vgg16.} Similar to results in Figure 3 in main text, each row represents a model with a different First Layer Kernel Size(FLKS), and each column corresponds to protected attribute groups. We observed a similar trending as discovered in the other 2 results above: generally, the perturbation shifts it's attention to low-frequency information as FLKS increases, and the perturbations for Black always have lower high-frequency focus compared to other race group.}
    \label{fig:spectra_vgg}
\end{figure*}

\begin{figure*}[!hbtp]
    \centering
    
    {\label{fig:1}\includegraphics[width=\textwidth]{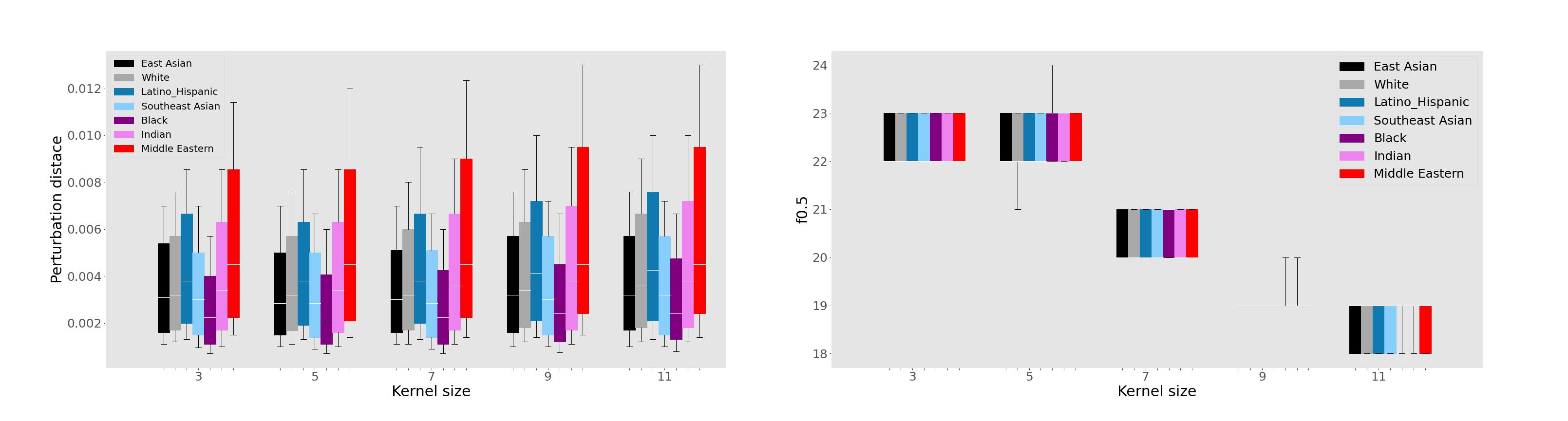}}
    \caption{\textbf{Perturbation spectra $f_{0.5}$ \& Perturbation distance for Vgg16.} We visualize the perturbation and $f_{0.5}$ scores in the same way discussed in Sec. 4.1 and Figure 4 We observed a similar trend: there is a significant trend that the  $f_{0.5}$ drops as the FLKS increases for all demographic groups and as the FLKS increases, the perturbation distances generally increase too for all the demographic groups.}
    \label{fig:f0.5_vgg}
\end{figure*}

To further test the robustness and universality of our framework and conclusion, we also tested on Vgg16-- another popular face analysis model. We also vary the first convolutional kernel size from $\{3,5,7,,9,11\}$. We report the averaged spectra in Figure \ref{fig:spectra_vgg}, and its corresponding perturbation distance \& $f_{0.5}$ scores in Figure \ref{fig:f0.5_vgg}.

\clearpage

\section{Regression results}
 We report the coefficients $\boldsymbol{\beta}$ (left) and $\boldsymbol{\gamma}$ (right) for the regression described in Section 4.1.3. We also report the standard deviations of the coefficient values and calculate their $t$ values according to $t = \frac{coef value}{std err}$, as well as $P > |t|$. A $P \leq 0.05$ indicates the value is significant. 

\begin{table*}[h]
  \caption{\textbf{Regression results of perturbation distance}. The $\mathbf{\beta}$ coefficient names use subscripts corresponding to race (E: East Asian, W: White, B: Black, I: Indian, L: Latino, S: Southeast Asian), and gender (M: Male, F: Female). Some large disparities between $\boldsymbol{\beta}$ values across groups are obvious, e.g., the values for Black Asian Female is $\sim 100\%$ higher than that of East Asian Female. Refer to Figure 6 in Main paper for plots on $\boldsymbol{\beta}$.}
  \label{tab:regress_dist}
  \begin{tabular}{cccrr | cccrr}
    \toprule
    coef name&coef value&std err&t&$P>|t|$&coef name&coef value&std err&t&$P>|t|$\\
    \midrule
    $\boldsymbol{\beta}_{EM}$&0.0227&0.003&6.930&0.000&$\boldsymbol{\gamma}_{EM}$&0.4251&0.025&17.308&0.000\\
    $\boldsymbol{\beta}_{EF}$&0.0274&0.003&8.323&0.000&$\boldsymbol{\gamma}_{EF}$&0.4232&0.025&17.065&0.000\\
    $\boldsymbol{\beta}_{WM}$&0.0306&0.003&11.392&0.000&$\boldsymbol{\gamma}_{WM}$&0.4453&0.020&21.969&0.000\\
    $\boldsymbol{\beta}_{WF}$&0.0254&0.003&8.614&0.000&$\boldsymbol{\gamma}_{WF}$&0.4302&0.022&19.361&0.000\\
    $\boldsymbol{\beta}_{LM}$&0.0351&0.003&11.143&0.000&$\boldsymbol{\gamma}_{LM}$&0.2088&0.012&17.566&0.000\\
    $\boldsymbol{\beta}_{LF}$&0.0269&0.003&8.521&0.000&$\boldsymbol{\gamma}_{LF}$&0.2425&0.012&20.304&0.000\\
    $\boldsymbol{\beta}_{SM}$&0.0251&0.003&7.552&0.000&$\boldsymbol{\gamma}_{BM}$&0.1856&0.013&14.783&0.000\\
    $\boldsymbol{\beta}_{SF}$&0.0189&0.004&5.982&0.000&$\boldsymbol{\gamma}_{BF}$&0.2163&0.014&15.281&0.000\\
    $\boldsymbol{\beta}_{BM}$&0.0589&0.001&48.770&0.000&$\boldsymbol{\gamma}_{BM}$&0.2088&0.012&17.505&0.000\\
    $\boldsymbol{\beta}_{BF}$&0.0659&0.001&48.588&0.000&$\boldsymbol{\gamma}_{BF}$&0.2425&0.012&20.234&0.000\\  
    $\boldsymbol{\beta}_{IM}$&0.0748&0.001&61.160&0.000&$\boldsymbol{\gamma}_{BM}$&0.1856&0.013&14.783&0.000\\
    $\boldsymbol{\beta}_{IF}$&0.0846&0.001&65.294&0.000&$\boldsymbol{\gamma}_{BF}$&0.2163&0.014&15.821&0.000\\  
    
  \bottomrule
\end{tabular}
\end{table*}

\clearpage

\section{Results of varying all convolutional kernel sizes}
\begin{figure*}[h]
    \centering
    
    {\label{fig:1}\includegraphics[width=0.8\textwidth]{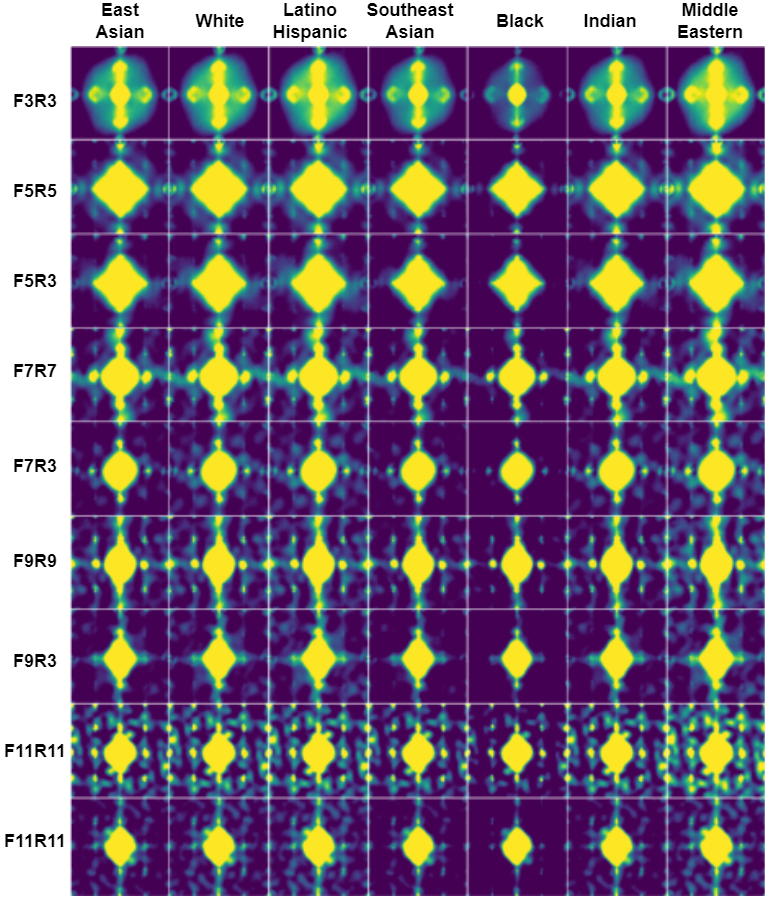}}
    \caption{\textbf{Perturbation Spectrum Visualization on Fairface for modifying all convolutional layer kernel size.} Each row represents a model with a different architectual choice, for example, ``F3R3'' represents model with first layer kernel size of 3 and the rest layers kernel sizes of 3, and each column corresponds to protected attribute groups. We found that modifying all convolutional layers' kernel sizes doesn't make a significant difference comparing to {\it only} modify the first convolutional kernel size.}
    \label{fig:spectra_all}
\end{figure*}

We also test our framework on the occasion where we modify all the convolutional layers' kernel sizes. The results are in Figure \ref{fig:spectra_all}. Basically, we found that modifying all convolutional layers' kernel sizes doesn't make a significant difference comparing to {\it only} modify the first convolutional kernel size. Refer to the caption for more details.

\clearpage
\section{Results of applying CW attack to model trained on UTKFace}
\begin{figure*}[h] 
    \centering  
    {\label{fig:1}\includegraphics[width=0.7\textwidth]{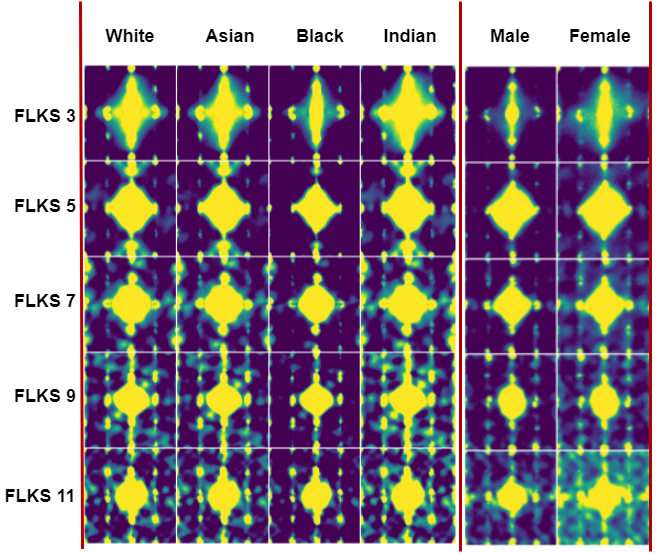}}

    \caption{\textbf{Average spectra of adversarial perturbation images split by race and gender for UTKFace}. We see similar trends in these spectra to the ones shown for Fairface in Fig 3 in Main paper.}
    \label{fig:spectra_utk}
\end{figure*}

We also conduct experiments on UTKFace, another popular face image dataset and report the results in Figure \ref{fig:spectra_utk}. Refer to Figure 3 in main text for results on Fairface and analysis.

\clearpage
\section{Results of applying FGSM to model}
We also test our framework on the occasion where we apply FGSM attack to all the models trained on Fairface. The results are in Figure \ref{fig:spectra_fgsm}. It has basically the same trend with all the previous spectra visualization.
\begin{figure*}[h]
    \centering
    
    {\label{fig:1}\includegraphics[width=\textwidth]{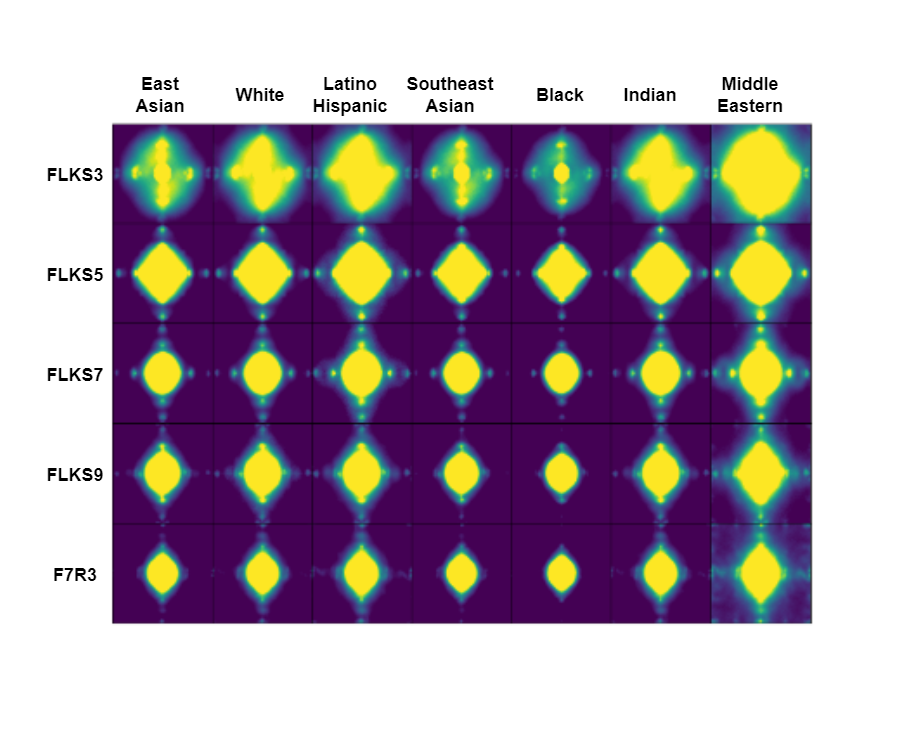}}
    \caption{\textbf{Perturbation Spectrum Visualization on Fairface using FGSM.} Similar to results in Figure 4, each row represents a model with a different First Layer Kernel Size(FLKS), and each column corresponds to protected attribute groups. We observed a similar trending: generally, the perturbation shifts it's attention to low-frequency information as FLKS increases, and the perturbations for Black always have lower high-frequency focus compared to other race group.}
    \label{fig:spectra_fgsm}
\end{figure*}

\clearpage
\section{Frequency energy injection result}
Same to Figure 5 in main paper, we show models’ performances with different FLKS for all race groups separately.

\begin{figure*}[h]
    \centering
    
    {\label{fig:1}\includegraphics[width=\textwidth]{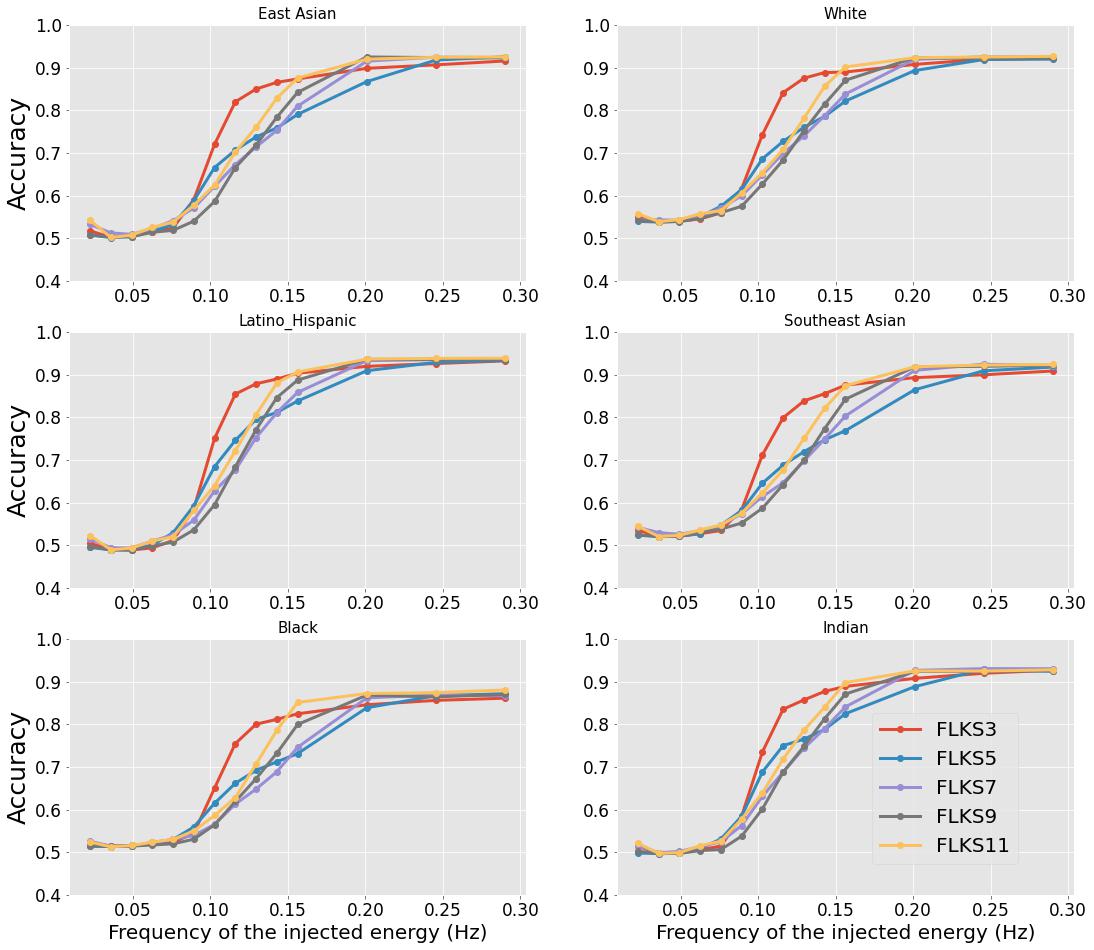}}
    \caption{\textbf{Frequency energy injection result.} In each individual figure, the x-axis is the frequency we are injecting energy at and the y-axis is the accuracy of different models. It is obvious that all the models suffer from low to mid frequency's energy injections, and become robust to mid to high frequency noises. It is hard to directly tell which group is getting influenced more than the others, which furthers asks for a quantitatively analysis.}
    \label{fig:spectra_fgsm}
\end{figure*}

\end{document}